\documentclass[11pt]{article}

\usepackage[preprint]{acl}

\usepackage{times}
\usepackage{latexsym}

\usepackage[T1]{fontenc}
\usepackage[utf8]{inputenc}

\usepackage{microtype}

\usepackage{graphicx}

\usepackage{hyperref}
\usepackage{url}
\usepackage{booktabs}
\usepackage{amsfonts}
\usepackage{amsmath}
\usepackage{amssymb}
\usepackage{nicefrac}
\usepackage{microtype}
\usepackage{xcolor}
\usepackage{graphicx}
\usepackage{pgfplots}
\pgfplotsset{compat=1.18}
\usepackage{float}
\usepackage[most]{tcolorbox}
\tcbset{
  algbox/.style={
    enhanced,
    colback=white,
    colframe=black,
    boxrule=0.8pt,
    arc=0pt,
    left=6pt, right=6pt, top=4pt, bottom=4pt,
    before upper={\setlength{\parskip}{2pt}},
  }
}
\usepackage{listings}
\makeatletter
\newcounter{algorithm}
\renewcommand{\thealgorithm}{\arabic{algorithm}}
\newfloat{algorithm}{tbp}{loa}
\floatname{algorithm}{Algorithm}
\newenvironment{algorithmic}[1][]{%
  \begin{list}{}{\leftmargin=1.5em\itemsep=2pt\parsep=0pt\topsep=4pt}%
}{%
  \end{list}%
}
\newcommand{\State}{\item}
\newcommand{\If}[1]{\item \textbf{if} #1 \textbf{then}}
\newcommand{\Else}{\item \textbf{else}}

\newcommand{\EndIf}{\item \textbf{end if}}
\newcommand{\While}[1]{\item \textbf{while} #1 \textbf{do}}
\newcommand{\EndWhile}{\item \textbf{end while}}
\newcommand{\For}[1]{\item \textbf{for} #1 \textbf{do}}
\newcommand{\EndFor}{\item \textbf{end for}}
\newcommand{\Return}{\item \textbf{return}~}
\newcommand{\Require}{\item \textbf{Require:}~}

\newcommand{\Comment}[1]{\hfill\textit{// #1}}
\makeatother
\usepackage{tikz}
\usetikzlibrary{arrows.meta, positioning, shapes.geometric, fit, calc}
\usepackage{multirow}
\usepackage{array}
\usepackage{wrapfig}
\usepackage{placeins}

\usepackage{cleveref}
\crefformat{section}{\S#2#1#3} % see manual of cleveref, section 8.2.1
\crefformat{subsection}{\S#2#1#3}
\crefformat{subsubsection}{\S#2#1#3}

\newcommand{\TRADE}{\textsc{TRADE}}
\newcommand{\TAED}{\textsc{TAED}}
\newcommand{\eg}{\emph{e.g.}}
\newcommand{\ie}{\emph{i.e.}}

\title{TRADE: Transducer-Augmented Decoder for Speech LLM}

\author{
  \textbf{Yun Tang\textsuperscript{1}},
  \textbf{Shanil Puri\textsuperscript{1}},
  \textbf{Shinji Watanabe\textsuperscript{2}},
  \textbf{Subhabrata Mukherjee\textsuperscript{1}}
\\
\\
  \textsuperscript{1}Hippocratic AI,
  \textsuperscript{2}Carnegie Mellon University,
\\
  \small{
    \href{mailto:yuntang.email@gmail.com}{yuntang.email@gmail.com},
    \href{mailto:shanil@hippocraticai.com}{shanil@hippocraticai.com},
  }
}

\begin{document}
\maketitle

\begin{abstract}
Speech Large Language Models (Speech LLMs) lack a principled mechanism for streaming inference: their label-synchronous generation has no acoustic-frame alignment, making real-time decoding and end-of-utterance detection difficult.
We propose \TRADE{} (\textbf{TR}ansducer-\textbf{A}ugmented \textbf{DE}coder), which augments a multimodal LLM with a transducer branch that shares the audio encoder and uses the LLM's hidden states directly as the prediction network --- coupling frame-synchronous acoustic alignment with the LLM's linguistic reasoning.
Three design choices make the system accurate, streamable, and long-form capable:
(1)~\textbf{Tightly coupled dual vocabularies} --- a compact transducer vocabulary derived from the LLM vocabulary, enabling zero-cost score fusion;
(2)~\textbf{Chunk-synchronized streaming training} with gradient stopping, eliminating the train--inference mismatch at offline-equivalent memory cost; and
(3)~\textbf{Localized Decoder Audio Attention (LDAA)}, a causal sliding window that caps KV-cache memory independently of utterance length.
A single \TRADE{} checkpoint supports offline and streaming decoding across a continuous range of latency operating points.
\TRADE{} achieves 6.71\% average WER on the Open ASR Leaderboard, 
while the streaming recognition with 960ms chunk size reaches 8.40\% from the same checkpoint. On long-form speech, it obtains 3.64\% WER on TED-LIUM and 10.88\% on Earnings-22 without external segmentation. \TRADE{} provides sentence-end punctuation timestamps that, when combined with acoustic voice activity detection (VAD), improve end-of-utterance detection by +0.03 $F_1$ over acoustic VAD alone.
\end{abstract}

% ===============================================================
\section{Introduction}
\label{sec:intro}

Speech Large Language Models (Speech LLMs) have emerged as a compelling paradigm for end-to-end speech understanding, leveraging the rich linguistic prior of pre-trained language models to achieve strong recognition and comprehension capabilities. A fundamental limitation, however, is that the LLM decoder is entirely \emph{label-synchronous} — it generates one token per autoregressive step with no explicit alignment to acoustic frames — making it ill-suited for streaming inference and lacking a principled mechanism for end-of-utterance detection~\citep{Arivazhagan2019MonotonicIL,xma2020monotonic}.

Existing approaches to streaming in Speech LLMs fall into two broad families.
The first covers \emph{hard-coded} streaming strategies, which include both fixed-chunk methods that feed each audio chunk as a prefix to the LLM input sequence~\citep{chen2024bestow, deng2025simuls2s} and interleaved token-stream models that mix speech and text tokens in a single unified sequence~\citep{defossez2024moshi, xie2024miniomni, nguyen2024spiritlm}.
Both variants are hard-coded in the same fundamental sense: the timing of token emission is governed by an externally fixed policy (chunk boundary or token interleaving schedule) rather than by any learned acoustic alignment, leaving the model with no principled mechanism to decide \emph{when} a particular token is grounded in the audio.
The second family augments the LLM decoder with a CTC or transducer auxiliary to impose frame-level supervision~\citep{watanabe2017hybrid, moriya2024allinone, seide2024reallm}: alignment guidance is provided as a side signal or via special time tokens, but the LLM autoregressive head remains the primary output mechanism and the coupling between acoustic timing and language generation stays loose.
Neither family provides tight acoustic--linguistic coupling: the LLM and the acoustic alignment mechanism remain largely separate, with no shared hidden state between them.

We propose \TRADE{} (\textbf{TR}ansducer-\textbf{A}ugmented \textbf{DE}coder), which extends the Hybrid \TAED{}~\citep{tang2023taed} framework to the decoder-only Speech LLM setting. \TRADE{} augments a multimodal LLM with a transducer branch that shares the audio encoder and uses the LLM's hidden states directly as the prediction network. The transducer is the primary decoder — it controls when to advance the acoustic frame or emit a token — while the LLM provides linguistic context and full vocabulary coverage at every step through score fusion. This tight coupling gives \TRADE{} frame-synchronous acoustic alignment without sacrificing the LLM's linguistic reasoning.

The key technical contributions are:
\begin{enumerate}
  \item \textbf{Joint transducer–LLM architecture.} \TRADE{} shares a single audio encoder across the LLM and transducer paths and uses the LLM's hidden states directly as the transducer prediction network, tightly coupling frame-synchronous acoustic alignment with autoregressive linguistic reasoning. (\cref{sec:overview})
  \item \textbf{Tightly coupled dual vocabularies.} A compact transducer vocabulary is \emph{derived} from the LLM vocabulary by preserving original token IDs and merging pronunciationally equivalent surface forms, making the transducer lattice tractable. At inference the two vocabularies actively collaborate: LLM probability mass is marginalized per homophone set and fused with transducer scores, recovering full surface-form quality — casing, punctuation, and spelling variants — without post-processing. (\cref{sec:vocab})
  \item \textbf{Memory-efficient streaming training.} Dynamic chunk-based synchronized training ties LLM re-prefill to chunk boundaries, eliminating the train--inference mismatch and enabling a single checkpoint to operate across a range of latency--accuracy trade-offs. Gradient stopping at the LLM boundary reduces peak memory to the same order as offline training. (\cref{sec:streaming_modes})
  \item \textbf{Localized Decoder Audio Attention.} A causal sliding window confines the LLM's audio attention to a bounded span of recent context. This caps the inference KV-cache regardless of utterance length, enabling long-form ASR without memory growth, and as a side benefit discards stale early-utterance context that can otherwise drift alignment. (\cref{sec:windowed})
\end{enumerate}

Experimentally, \TRADE{} achieves competitive recognition accuracy on the Open ASR Leaderboard benchmark; supports seamless streaming across a range of latency operating points from a single checkpoint; transcribes long-form audio natively — without VAD segmentation or chunked batching — through its frame-synchronous streaming decoder; and improves utterance boundary detection by fusing the transducer's punctuation emissions with acoustic voice-activity signals.

% ===============================================================
\section{Background}
\label{sec:background}

\paragraph{Transducer.}
The RNN-T~\citep{graves2012sequence} factorizes output prediction into an encoder, a prediction network, and a joint network. A blank symbol acts as a read gate: the model either emits a token or advances the acoustic frame, yielding a frame-synchronous, inherently streaming alignment trained end-to-end via the transducer loss.

\paragraph{Hybrid \TAED{} Architecture.}
The Hybrid Transducer and Attention-based Encoder–Decoder (\TAED{})~\citep{tang2023taed} combines transducer streaming with an AED decoder under a shared encoder, jointly optimized with transducer and cross-entropy losses. The AED decoder's hidden states serve directly as the transducer prediction network. 
The tight coupling is shown to boost accuracy over a purely acoustic transducer significantly~\citep{tang2025enhanced}.

\paragraph{Chunk-synchronized training.}
Standard streaming training suffers a train–inference mismatch because the decoder state $s_u(t)$ is computed from incomplete encoder context, where $u$ is the last decoded token and $t$ is timestamp of available audio input. \TAED{} resolves this by refreshing the decoder state at each chunk boundary $\delta(t)$:
\begin{equation}
  s_u(t) = f_{\mathrm{dec}}\!\left(h_{1:\delta(t)},\; y_{<u}\right),
  \label{eq:chunk_sync}
\end{equation}
where $h_{1:\delta(t)}$ are encoder states up to the boundary and $y_{<u}$ are previously decoded tokens. %Setting chunk size to the full utterance reduces the scheme to standard offline training.

% ===============================================================
\section{\TRADE{} Model}
\label{sec:model}

\subsection{Overview}
\label{sec:overview}

\begin{figure}[t]
\centering
\includegraphics[width=\columnwidth]{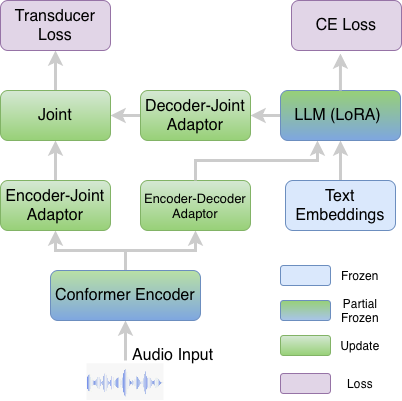}
\caption{\TRADE{} architecture. A shared Conformer encoder feeds both an \emph{LLM path} (cross-entropy loss) and a \emph{transducer path} (transducer loss); the LLM hidden states serve as the transducer prediction network via the Decoder-to-Joint Adaptor.}
\label{fig:dat}
\end{figure}

\TRADE{} augments a multimodal LLM with a transducer branch, as illustrated in Figure~\ref{fig:dat}. An LLM can be viewed as a variant of the AED framework in which encoder and decoder are unified within a single large transformer.  
\TRADE{} extends the core principles of \TAED{} to this setting, including joint transducer and cross-entropy training, a shared encoder, decoder-conditioned prediction states, and chunk-synchronized training with a full pre-trained LLM decoder.

A shared Conformer encoder feeds two parallel paths. In the \emph{LLM path}, encoder outputs are projected into the LLM embedding space through an adaptor, producing \emph{LLM audio embeddings}. These embeddings are concatenated with text token embeddings and processed by the LLM under a cross-entropy objective. In the \emph{transducer path}, a separate adaptor projects acoustic features into the transducer joint network, while the LLM final hidden states are projected into the prediction space through a lightweight linear adaptor, replacing the conventional RNN-based prediction network.

The joint network combines acoustic and prediction features at every $(t,u)$ lattice point and is trained with a transducer loss. The two paths share the encoder and are optimized jointly. Sections~\ref{sec:encoder}--\ref{sec:predictor} describe each component; the training objective is in Section~\ref{sec:objective}.

\subsection{Encoder}
\label{sec:encoder}

The shared acoustic backbone is a Conformer encoder~\citep{gulati2020conformer,Rekesh2023FastCW}, with the top layers fine-tuned and the remainder frozen. SpecAugment~\citep{park2019specaugment} is applied during training. For streaming, we adopt the Copy-and-Append Data Augmentation (CADA) scheme~\citep{liu2021crossattention, tang2025chunk}, which exposes exactly one lookahead chunk per layer without cascading future context; the same encoder checkpoint serves both offline and streaming inference by employing different chunk sizes. Full details are in Appendix~\ref{app:cada}.

\subsection{LLM as Prediction Network with Dual Vocabularies}
\label{sec:predictor}

The LLM serves a dual role: it generates the transcription under cross-entropy supervision and supplies prediction states to the transducer joint network. After each LLM forward pass, a lightweight linear adaptor gathers the LLM's last hidden states at positions used to predict verbalized tokens and projects them into the joint network's prediction space. % A learned start state prepended before the first output token ensures proper transducer alignment.
The joint network combines acoustic features $h_t$ from the encoder adaptor and prediction features $s_u(t)$ from the LLM at each $(t, u)$ lattice point, producing a distribution over the compact transducer vocabulary $\mathcal{V}_\text{trans}$ through a standard additive joint. Unlike \TAED{}~\citep{tang2023taed}, which uses a dedicated AED decoder as the prediction network, \TRADE{} supplies $s_u(t)$ directly from LLM hidden states.

\begin{figure*}[t]
\centering
\small
\begin{tcolorbox}[algbox]
\textbf{Transcription:}\quad\textit{``Yes, I'm OK. It's great!''}\\[5pt]
\renewcommand{\arraystretch}{1.5}\setlength{\tabcolsep}{4pt}
\begin{tabular}{@{}r*{9}{>{\centering\arraybackslash\ttfamily\small}p{1.1cm}}@{}}
\normalfont\textbf{Tokenized:}  & Yes & {,} & I'm & OK& {.} & It & 's & great & {!} \\[2pt]
\normalfont\textbf{Verbalized:} & \textcolor{teal}{yes} & \textcolor{gray!80}{---} & \textcolor{teal}{i'm} & \textcolor{teal}{ok} & \textcolor{gray!80}{---} & \textcolor{teal}{it} & \textcolor{teal}{'s} & \textcolor{teal}{great} & \textcolor{gray!80}{---} \\
\end{tabular}
\end{tcolorbox}
\caption{Comparison of LLM tokens and verbalized tokens.
\textcolor{teal}{\textbf{Verbalized token}} map to the compact transducer vocabulary; predicted by the joint network.
\textcolor{gray!70}{\textbf{Non-verbalized token}} correspond to punctuation and formatting tokens absent from the transducer lattice; emitted by the LLM at blank steps.}
\label{fig:vocab_example}
\end{figure*}
\paragraph{Dual Vocabularies: construction.}\label{sec:vocab}

\TRADE{} operates over two \emph{tightly coupled} vocabularies: a full LLM vocabulary ($|\mathcal{V}_\text{llm}|$ ${\approx}128$K tokens for Llama-3) and a compact transducer vocabulary ($|\mathcal{V}_\text{trans}|$ ${\approx}20$K tokens).
The coupling is twofold. First, $\mathcal{V}_\text{trans}$ is derived directly from $\mathcal{V}_\text{llm}$. Second, both vocabularies participate jointly during decoding: the transducer performs acoustic alignment in the compact space, while the LLM recovers surface forms and non-verbalized tokens (Section~\ref{sec:formatted}).

A transducer must anchor each emission to an acoustic frame, ruling out non-verbalized tokens such as punctuation and whitespace. Moreover, the $T \times U \times V$ transducer lattice becomes prohibitively expensive when using the full 128K-token vocabulary, where $T$, $U$, and $V$ denote the number of acoustic frames, output tokens, and vocabulary size, respectively.

Vocabulary construction proceeds in two stages (Appendix~\ref{app:vocab_construction}): 
(1) a \emph{pruned tokenizer} retains only acoustically realizable tokens while \emph{preserving original LLM token IDs}, enabling zero-cost joint decoding; and
(2) \emph{surface-form normalization} merges homophone variants (\eg{} ``\textit{OK}'' / ``\textit{ok}'') into canonical transducer token IDs.

\paragraph{Dual Vocabularies: inference-time collaboration.}\label{sec:formatted}

During inference, the two vocabularies operate jointly, as illustrated in Figure~\ref{fig:vocab_example}. Shared token IDs provide the bridge between the vocabularies, allowing LLM scores over $\mathcal{V}_\text{llm}$ to be mapped efficiently into $\mathcal{V}_\text{trans}$.

\textbf{a) Verbalized token selection.} On a non-blank transducer emission, let $\log p^\text{trans}_c$ and $\log p^\text{LLM}_v$ denote the normalized log-probabilities (log-softmax) of token $c$ in the transducer vocabulary and $v$ in LLM vocabulary. The LLM scores are projected into compact space by summing probability mass over all LLM tokens associated with each transducer token $c$:
\begin{equation}
  \log \tilde{p}^\text{LLM}_c = \operatorname{logsumexp}_{v \in \mathcal{H}(c)} \log p^\text{LLM}_v,
  \label{eq:lm_marginalize}
\end{equation}
where $\mathcal{H}(c)$ denotes the homophone set associated with $c$. The transducer and LLM scores are then fused in the compact vocabulary space:
\begin{equation}
  \hat{c} = \arg\max_{c \in \mathcal{V}_\text{trans} \setminus \{\varnothing\}}
    \bigl[\, w \log p^\text{trans}_c + (1-w) \log \tilde{p}^\text{LLM}_c \,\bigr],
  \label{eq:fusion}
\end{equation}
with fusion weight $w$ (default $w=0.5$).  If multiple homophones exist for $\hat{c}$, the final surface form is selected using
$\hat{v} = \arg\max_{v \in \mathcal{H}(\hat{c})} \log p^\text{LLM}_v$.
%This keeps alignment in compact space while recovering the correct surface form via the LLM (e.g.\ \textit{yes}$\to$\textit{Yes}, \textit{it}$\to$\textit{It} in Figure~\ref{fig:vocab_example}).
This design preserves acoustic alignment in the compact vocabulary while allowing the LLM to recover the appropriate surface form (e.g., \textit{yes}$\rightarrow$\textit{Yes}, \textit{it}$\rightarrow$\textit{It} in Figure~\ref{fig:vocab_example}).

\textbf{b) Non-verbalized token recovery.} On a blank emission (read step), before advancing the acoustic frame the LLM is queried for its next-token prediction; any leading non-verbalized tokens, such as punctuations (gray entries in Figure~\ref{fig:vocab_example}), are emitted immediately. The ``\textit{,}'' after \textit{Yes} and the ``\textit{.}'' after \textit{OK} are recovered this way, producing the fully punctuated output \textit{``Yes, I'm OK.\ It's great!''} without any post-processing.

% ===============================================================
\section{Training and Inference}
\label{sec:streaming}

\subsection{Training Objective}
\label{sec:objective}

The total training loss combines the LLM cross-entropy and the transducer loss:
\begin{equation}
  \mathcal{L}_\text{total} = (1 - \alpha)\,\mathcal{L}_\text{ce} + \alpha\,\mathcal{L}_\text{trans},
  \label{eq:loss}
\end{equation}
with $\alpha = 0.5$ as default. The transducer loss $\mathcal{L}_\text{trans}$ is computed with the k2 pruned RNNT algorithm~\citep{kuang2022pruned}, which restricts the lattice to gradient-estimated high-probability arcs, making large-vocabulary transducer training tractable.
During training, the LLM operates in teacher-forcing mode over the full token sequence.
For each verbalized token $u$, the prediction feature is taken from the LLM hidden state immediately preceding $u$, which may correspond to a non-verbalized token. For example, in Figure~\ref{fig:vocab_example}, the prediction feature for ``\textit{It}'' is the hidden state at the period token ``\texttt{.}'' (non-verbalized), since the LLM predicts ``\textit{It}'' immediately after ``\texttt{.}''. Consequently, we use the hidden state of ``\texttt{.}'' rather than ``\textit{OK}'' as input to the joint network.

\subsection{Localized Decoder Audio Attention}
\label{sec:windowed}

Without windowing, the LLM's audio context $h_{1:\delta(t)}$ grows linearly with utterance length, causing memory usage and latency to increase substantially for long-form ASR. To address this, we propose \textbf{Localized Decoder Audio Attention (LDAA)}, which constrains the LLM audio context to a bounded sliding window during both training and inference. This design provides two key benefits:
%making long-form ASR increasingly expensive in memory and latency. We propose \textbf{Localized Decoder Audio Attention (LDAA)}, which constrains the LLM's audio context to a bounded sliding window applied during \emph{both} training and inference. This yields two concrete benefits:
(1) \emph{constant-memory streaming} — KV-cache size is capped at $(N_l+2) \cdot C$ LLM audio embedding frames (i.e., $N_l$ left-context chunks, one current chunk, and one lookahead chunk),  regardless of utterance length; and (2) \emph{shorter, more focused attention} — the LLM attends only to the acoustically relevant neighbourhood of the current decoding position, reducing prefill latency and avoiding stale context from earlier in the utterance.

\paragraph{Sliding window formulation.}
At acoustic frame $t$, the LLM observes the audio interval $[\tau^-_{\delta(t)},\, \tau^+_{\delta(t)})$, where $\tau^-_{\delta(t)} = \max\big(0,\, \delta(t) - (N_l+1) \cdot C\big)$ and $\tau^+_{\delta(t)} = \delta(t) + C$.
The visible context therefore yields a sliding window of at most $(N_l+2)\cdot C$ frames. During the initial stage of an utterance, the window grows monotonically. Once $t \ge N_l \cdot C$, the window advances chunk by chunk, discarding frames older than $N_l$ chunks.

\paragraph{Window size selection.}
The left-context duration must cover the acoustic evidence needed to emit each token reliably. We quantify this via acoustic support span analysis (Appendix~\ref{app:emission}): the 95th-percentile span is 2.56\,s and the 99th-percentile is 3.28\,s. In this study, we choose 5\,s as default left-context duration, which provides a comfortable margin above 99th-percentile 3.28\,s.

\subsection{Chunk-based Synchronized Training}
\label{sec:streaming_modes}

\TRADE{} extends the chunk-synchronized training scheme introduced in \TAED{} (Eq.~\ref{eq:chunk_sync}) to the LLM setting. Compared with \TAED{}, 
the decoder state is instantiated through an LLM forward pass refreshed at each chunk boundary under the LDAA constraint:
\begin{equation}
    s_u(t) = f_{\mathrm{llm}}\!\left(h_{\tau^-_{\delta(t)}:\tau^+_{\delta(t)}},\; y_{<u}\right).
    \label{eq:llm_sync}
\end{equation}

\paragraph{Dynamic chunk training.}
We apply dynamic chunk training~\citep{Zhang2020UnifiedSA,Weninger2022ConformerWD} to \TRADE{}, 
where the chunk size $C$ at each training step is randomly sampled from a predefined set of candidate sizes. The candidates range from short chunks that simulate low-latency streaming conditions to full-context training, which reduces to standard offline training (see Appendix~\ref{app:config} for the exact sampling distribution).
This multi-granularity exposure trains the model to operate robustly across the full range of chunk sizes, allowing a single checkpoint to be deployed at different latency-accuracy operating points without retraining.

\paragraph{Gradient stopping from transducer to LLM.}
In streaming mode, the LLM state $s_u(t)$ in Eq.~(\ref{eq:llm_sync}) is recomputed at every chunk boundary $\delta(t)$. As a result, the transducer backward pass would otherwise need to retain a separate LLM activation graph for each chunk, causing memory usage to grow linearly with the number of chunks in the utterance.

To avoid this overhead, we stop gradients at the LLM boundary during streaming training: hidden states passed to the joint network are detached, so no LLM activations are retained for the transducer backward pass. For full-context training steps, only a single LLM forward pass is required, and gradients flow normally. This strategy reduces peak memory usage to the same order as offline training.

\begin{algorithm}[!t]
\caption{Streaming fused decoding with localized decoder audio attention}
\label{alg:stream}
\scriptsize
\begin{tcolorbox}[algbox]
\begin{algorithmic}[1]
\Require Audio stream (chunks $k = 0, 1, \ldots$), text prompt embedding $\mathbf{x}$, fusion weight $w$, left-context $N_l$ chunks
\State $\mathbf{y} \gets [\,]$ \Comment{Accumulated transcript}
\For{each arriving chunk $k+1$}
 \State Encode chunk $k$ and chunk $k+1$ (lookahead) using the streaming audio encoder
  
  \State Compute the windowed interval $[\tau^-_{\delta(t)}, \tau^+_{\delta(t)})$ as defined in Section~\ref{sec:windowed}
  
  \State Re-prefill the LLM with $\mathbf{x}$, windowed audio embeddings $h_{\tau^-_{\delta(t)}:\tau^+_{\delta(t)}}$, and $\mathbf{y}$
  \State \hspace{1.5em}$\Rightarrow$ obtain hidden state $\mathbf{s}$ and output logit $\boldsymbol{\ell}$
  \For{each acoustic frame $t$ in chunk $k$}
    \State $\mathbf{z} \gets \mathrm{Joint}(h_t, \mathbf{s})$;\quad $\hat{v} \gets \arg\max\,\mathbf{z}$
    \If{$\hat{v} = \varnothing$ (blank)}  \Comment{Read}
    
      \While{top LLM token is non-verbalized}
      
        \State Emit top LLM token; advance LLM autoregressively; update  $\mathbf{s}$ and $\boldsymbol{\ell}$
      \EndWhile
    \Else  \Comment{Write}
      \State $\log p^\text{trans} \gets \log\mathrm{softmax}(z)$;\quad $\log p^\text{llm} \gets \log\mathrm{softmax}(\boldsymbol{\ell})$
      
      \State $\log \tilde{p}^\text{llm}_c \gets \operatorname{logsumexp}_{v \in \mathcal{H}(c)} \log p^\text{llm}_v$
      \Comment{Marginalize the LLM distribution into $\mathcal{V}_\text{trans}$}
      
      \State $\hat{c} \gets \arg\max_{c \neq \varnothing}\!\left[w \log p^\text{trans}_c + (1-w) \log \tilde{p}^\text{llm}_c\right]$
      
      \State $\hat{y} \gets \arg\max_{v \in \mathcal{H}(\hat{c})} \log p^\text{llm}_v$  \Comment{Homophone disambiguation}

      \State Append $\hat{y}$ to $\mathbf{y}$; advance the LLM autoregressively; update  $\mathbf{s}$ and $\boldsymbol{\ell}$
    \EndIf
  \EndFor
  \State Release audio frames older than $\tau^-_{{\delta(t)}+1}$ from memory \Comment{Sliding window cleanup}
\EndFor
\State \Return $\mathbf{y}$
\end{algorithmic}
\end{tcolorbox}
\end{algorithm}

\subsection{Decoding}
\label{sec:decoding}

%\TRADE{} streaming inference is described in (Algorithm~\ref{alg:stream}). The transducer controls the read/write decision at each acoustic frame: on a non-blank emission, the transducer and LLM jointly select a verbalized token via score fusion (Eq.~\ref{eq:fusion}); on a blank emission, the LLM recovers any non-verbalized tokens before advancing the frame. 
Streaming inference for \TRADE{} is summarized in Algorithm~\ref{alg:stream}. At each acoustic frame, the transducer determines the read/write decision. For non-blank emissions, the transducer and LLM jointly select a verbalized token through score fusion (Eq.~\ref{eq:fusion}). For blank emissions, the LLM recovers any non-verbalized tokens before advancing to the next frame.

In streaming mode, audio arrives incrementally in chunks. At each chunk boundary, the LLM is re-prefilled using the prompt, the windowed audio embeddings $h_{\tau^-_{\delta(t)}:\tau^+_{\delta(t)}}$, and the current partial transcript.

This ensures bounded memory usage, with at most $(N_l+2)\cdot C$ LLM audio embedding frames retained in context at any time.
Offline inference is treated as a special case of streaming inference in which the chunk size equals the full utterance length $T$. In this setting, the LLM is prefixed only once, and the inner frame loop runs over all $T$ frames without re-prefilling.

% ===============================================================
\section{Related Work}
\label{sec:related}

\paragraph{Speech LLMs via encoder-adapter-LLM.}
The dominant paradigm connects a pretrained speech encoder to a frozen or lightly fine-tuned LLM via a learned adapter~\citep{wang2023slm,chen2023salm,ma2024embarrassingly}; subsequent models scaled to richer capabilities and larger data (SALMONN~\citep{tang2024salmonn} with dual encoders and a Q-Former, Seed-ASR~\citep{bai2024seedasr} with an MoE backbone trained on 20M hours), and decoder-only variants~\citep{gupta2024decoder} showed LLM decoders are competitive for ASR.
All of these delegate decoding to the LLM's autoregressive head with no frame-level alignment mechanism.

\paragraph{Streaming speech LLMs.}
Speech~ReaLLM~\citep{seide2024reallm} introduced special time tokens to impose real-time flow on a decoder-only LLM; BESTOW~\citep{chen2024bestow} combined prompt prepending with per-layer cross-attention to unify offline and streaming; Moshi~\citep{defossez2024moshi} and Mini-Omni~\citep{xie2024miniomni} interleave speech and text tokens for full-duplex interaction.
These approaches hard-code emission timing via chunk boundaries or token interleaving, or rely on loose auxiliary signals --- none provide a principled frame-synchronous alignment mechanism integrated with the LLM's hidden state.

\paragraph{\TRADE{} in context.}
Unlike the encoder-adapter-LLM family, \TRADE{} retains a time-synchronous transducer that directly controls frame consumption, providing principled end-pointing and avoiding the hallucination and repetition artifacts common in attention-only streaming models on long audio.
The LLM contributes score fusion and linguistic context rather than driving the decoding loop; gradient stopping (Section~\ref{sec:streaming_modes}) addresses the training-time coupling specific to this joint architecture.

% ===============================================================
\section{Experiments}
\label{sec:experiments}

\subsection{Setup}

\paragraph{Experimental configuration.}
The encoder is a FastConformer-XL initialized from Parakeet-TDT-0.6B-v2~\citep{shah2025canary} with the top six layers fine-tuned; the LLM is Llama-3.2-1B~\citep{grattafiori2024llama} fine-tuned with LoRA~\citep{hu2021lora}; the transducer operates over a 20K compact vocabulary derived from the LLM vocabulary.
All models are optimized with AdamW under a cosine annealing schedule, with dynamic chunk-size training for streaming robustness.
Word error rate (WER) is evaluated on Whisper-normalized text applied to both hypothesis and reference.
We train on a large multi-domain corpus of approximately 153{,}000 hours (see Appendix~\ref{app:data} for details) on 16$\times$H200 GPUs in two phases: Phase~1 trains for 20{,}000 steps on the full corpus; Phase~2 fine-tunes for 25{,}000 steps with a contextual ASR objective~\citep{Lakomkin2023EndtoEndSR}.
The final checkpoint is used for both leaderboard and long-form evaluation.
Full configuration details are given in Appendix~\ref{app:config}.

\subsection{Open ASR Leaderboard Evaluation}
\label{sec:leaderboard}

Table~\ref{tab:leaderboard} presents the main \TRADE{} results on the Open ASR Leaderboard English benchmark~\citep{srivastav2025openasr}.
We report WER (\%) on the eight test sets used by the leaderboard.
We include Whisper-large-v3~\citep{radford2023robust}, Parakeet-TDT-0.6B-v3 and Canary-1B-v2 results~\citep{shah2025canary} for reference purpose. 
%\TRADE{} shares Parakeet's encoder backbone and Canary's 1B decoder scale.
\emph{Decoder-only LLM} is our internal baseline: the same shared encoder feeding the same LLM, but trained with cross-entropy loss only (no transducer branch), i.e.\ a standard decoder-based speech LLM~\citep{gupta2024decoder}.
\emph{\TRADE{}} uses joint transducer--LLM decoding (Eq.~\ref{eq:fusion}); \emph{\TRADE{} (stream-960\,ms)} and \emph{\TRADE{} (stream-640\,ms)} are results from streaming mode with chunk sizes  960\,ms and 640\,ms respectively. Note, the literature models in Table~\ref{tab:leaderboard} are trained under different condition and datasets.  \emph{Decoder-only LLM} is closest apples-to-apples comparison with \emph{\TRADE{}}.

\begin{table*}[t]
\centering
\small
\setlength{\tabcolsep}{4.5pt}
\caption{WER (\%) on the Open ASR Leaderboard English benchmark~\citep{srivastav2025openasr}.
\emph{Decoder-only LLM} is our in-house decoder-based speech LLM baseline, trained with cross-entropy loss only (no transducer branch).
\emph{\TRADE{}} uses joint transducer--LLM decoding.
\emph{\TRADE{} (stream-960\,ms)} and \emph{\TRADE{} (stream-640\,ms)} are results from streaming mode with chunk sizes 960\,ms and 640\,ms respectively.
$\dagger$~average WER over the eight sets.}
\label{tab:leaderboard}
\begin{tabular}{lrcccccccc}
\toprule
\textbf{System} & \textbf{Avg}$^\dagger$ & \textbf{AMI} & \textbf{E22} & \textbf{Giga} & \textbf{LS-c} & \textbf{LS-o} & \textbf{SPGI} & \textbf{TED} & \textbf{Vox} \\
\midrule
Whisper-large-v3~\citep{radford2023robust}          &  7.44 & 15.95 & 11.29 & 10.02 & 2.01 & 3.91 & 2.94 & 3.86 &  9.54 \\
Parakeet-TDT-0.6B-v3~\citep{shah2025canary}         &  6.32 & 11.39 & 11.19 &  9.57 & 1.92 & 3.59 & 3.98 & 2.80 &  6.09 \\
Canary-1B-v2~\citep{shah2025canary}                 &  7.15 & 16.01 & 11.79 & 10.82 & 2.18 & 3.56 & 2.28 & 4.29 &  6.25 \\
\midrule
%Decoder-only LLM (ours)                             &  6.43 & 13.67 & 10.88 & 9.85 & 1.68 & 3.26 & 2.14 & 3.64 &  6.34 \\
Decoder-only LLM (ours)                             &  6.87 & 16.16 & 11.51 & 10.07 & 1.70 & 3.01 & 2.23 & 3.71 &  6.59 \\  % this is a dedicated Decoder only model, not from TRADE decoder mode
\midrule
\TRADE{} (ours)                            	    &  6.71 & 14.85 & 11.02 & 10.24 & 1.60 & 3.13 & 2.36 & 3.84 &  6.60 \\
\midrule
\TRADE{} (stream-960\,ms) (ours)                      &  8.40 & 17.16 & 15.62 & 11.07 & 2.00 & 4.07 & 4.42 & 4.61 &  8.22 \\
\TRADE{} (stream-640\,ms) (ours)                      &  9.35 & 18.04 & 16.23 & 11.25 & 2.29 & 5.00 & 4.60 & 4.98 &  9.35 \\
\bottomrule
\end{tabular}
\end{table*}

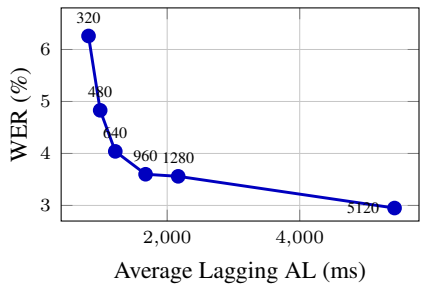
\begin{figure}[h]
\centering
\begin{tikzpicture}
\begin{axis}[
    width=0.82\columnwidth,
    height=4.4cm,
    xlabel={Average Lagging AL (ms)},
    ylabel={WER (\%)},
    xmin=400, xmax=5800,
    ymin=2.7, ymax=6.8,
    grid=both,
    grid style={line width=0.2pt, draw=gray!25},
    major grid style={line width=0.35pt, draw=gray!45},
    tick label style={font=\scriptsize},
    label style={font=\small},
    xlabel style={yshift=2pt},
    ylabel style={yshift=-4pt},
    clip=false,
]
\addplot[
    color=blue!75!black,
    mark=*,
    mark size=2.2pt,
    line width=1.1pt,
] coordinates {
    (813,  6.26)
    (989,  4.83)
    (1217, 4.04)
    (1674, 3.60)
    (2166, 3.56)
    (5431, 2.95)
};
\node[font=\tiny, above=1pt] at (axis cs:813,  6.26) {320};
\node[font=\tiny, above=1pt] at (axis cs:989,  4.83) {480};
\node[font=\tiny, above=1pt] at (axis cs:1217, 4.04) {640};
\node[font=\tiny, above=1pt] at (axis cs:1674, 3.60) {960};
\node[font=\tiny, above=1pt] at (axis cs:2166, 3.56) {1280};
\node[font=\tiny, left=2pt]  at (axis cs:5431, 2.95) {5120};
\end{axis}
\end{tikzpicture}
\caption{WER vs.\ Average Lagging (AL)~\citep{ma2019stacl} trade-off on \textbf{LibriSpeech} \textit{dev-other} across six chunk sizes (labels in ms). AL measures how much later each token is emitted relative to an ideal same-pace policy; lower AL indicates lower latency. } %\TRADE{} spans from aggressive 320\,ms streaming (WER\,=\,6.26\,\%, AL\,=\,813\,ms) to near-offline 5{,}120\,ms chunks (WER\,=\,2.95\,\%, AL\,=\,5{,}431\,ms) with a single checkpoint.}
\label{fig:wer_al}
\end{figure}

Crucially, all three operating points --- \TRADE{}, \TRADE{} (stream-960\,ms), and \TRADE{} (stream-640\,ms) --- are served by a \textbf{single checkpoint} with no architectural modification; the decoding mode is selected at inference time.
\TRADE{} achieves 6.71\% average WER, outperforming our cross-entropy-only decoder-based speech-LLM baseline (6.87\%) --- and doing so while supporting streaming inference, which the latter does not.
Moving to streaming, \TRADE{} (stream-960\,ms) reaches 8.40\% ($+1.69$ over offline) and \TRADE{} (stream-640\,ms) reaches 9.35\% ($+2.64$), enabling continuous low-latency transcription from the same single model.
Figure~\ref{fig:wer_al} shows the WER--latency trade-off across six chunk sizes on LibriSpeech \texttt{dev-other}; a full analysis including AL, DAL, AP, and RTF is in Appendix~\ref{app:latency}.

\subsection{Long-Form ASR}
\label{sec:longform}

Table~\ref{tab:longform} reports WER (\%) on three long-form benchmarks using \TRADE{} in streaming mode with 5{,}120\,ms chunks.
Beyond LDAA, we apply a 20 \emph{token sliding window}: the LLM conditions on at most the 20 most recently decoded tokens, with older tokens shifted to a buffer.
This keeps the KV-cache bounded and prevents the LLM from accumulating stale context that can cause repetition or hallucination over extended recordings.
Full decoding details are in Appendix~\ref{app:longform_decoding}. 

\begin{table}[t]
\small
\centering
\setlength{\tabcolsep}{4pt}
\caption{Long-form ASR WER (\%) on TED-LIUM~\citep{hernandez2018tedlium3}, Earnings-21~\citep{delrio2021earnings21}, and Earnings-22~\citep{delrio2022earnings22}.
\TRADE{} uses streaming decode with 5{,}120\,ms chunks.
$^{a}$Fast Conformer FT+LCA+GT~\citep{koluguri2024longform};
$^{b}$Canary-1B-v2 with parallel chunks~\citep{shah2025canary}.}
\label{tab:longform}
\begin{tabular}{lccc}
\toprule
\textbf{System} & \textbf{TED} & \textbf{E21} & \textbf{E22} \\
\midrule
FastConf.$^{a}$        &  4.98 & 13.84 & 19.49 \\
Canary-1B-v2$^{b}$     &    —  &    —  & 13.93 \\
\midrule
\TRADE{} (ours)          & \textbf{3.64} & \textbf{9.75} & \textbf{10.88} \\
\bottomrule
\end{tabular}
\end{table}

Unlike VAD-segmentation pipelines~\citep{koluguri2024longform} or parallel-chunk approaches~\citep{shah2025canary}, \TRADE{} requires no external segmentation: the transducer head drives streaming decoding within one pass, and the localized decoder audio attention window and token sliding window keep the LLM's context bounded regardless of recording length.
The GPU memory usage is around 8 Gigabytes during inference time.

%A detailed emission timing analysis — acoustic support span calibration for the LDAA window and end-of-utterance latency — is provided in Appendix~\ref{app:emission}.

% ===============================================================
\subsection{Ablation Study}
\label{sec:ablation}

\paragraph{Fusion weight sensitivity.}
Table~\ref{tab:fusion_sweep} sweeps $w \in \{0.1,\ldots,0.9\}$ on three test sets (\textbf{Librispeech} \textit{test-other}, \textbf{TED} and \textbf{Vox}).
The mean WER is flat within $0.06$\% absolute across $w \in [0.1, 0.7]$, with $w\!=\!0.3$ as the optimum (4.49\%); both endpoints lose $\sim$0.1\% absolute (decoder $w\!=\!0$: 4.58\%, transducer $w\!\to\!1$: 4.59\%).
Joint fusion outperforming both endpoints --- and the transducer endpoint nearly matching decoder mode despite its compact vocabulary --- indicates the two heads contribute complementary information: the transducer adds acoustic signal the LLM misses, while LLM marginalization (Eq.~\ref{eq:lm_marginalize}) recovers surface form the transducer's compact vocabulary cannot represent.

\paragraph{Vocabulary size.}
Table~\ref{tab:vocab_fused} ablates the compact transducer vocabulary size $|\mathcal{V}_\text{trans}| \in \{10{,}000, 15{,}000, 20{,}000\}$ across three decoding modes on the 8-set Open ASR Leaderboard.
$|\mathcal{V}_\text{trans}|\!=\!20{,}000$ matches or improves on smaller vocabularies in both offline and streaming modes. % every mode, with the largest gain under streaming ($-1.71$ abs WER vs 10K, $-2.49$ vs 15K).
We adopt $|\mathcal{V}_\text{trans}|\!=\!20{,}000$ as the default; per-testset breakdowns are in Appendix~\ref{app:vocab_ablation}.

\begin{table}[!htbp]
\centering
\small
\setlength{\tabcolsep}{3pt}
\caption{WER (\%) vs.\ fusion weight $w$ on three Open ASR Leaderboard test sets.
$\dagger$~mean over the five sets shown. Bold = per-column minimum.}
\label{tab:fusion_sweep}
\resizebox{\columnwidth}{!}{%
\begin{tabular}{lrrrr}
\toprule
\textbf{Mode / $w$}   & \textbf{LS-o} & \textbf{TED} & \textbf{Vox} & \textbf{Mean}$^\dagger$ \\
\midrule
Decoder ($w\!=\!0$)                              & 3.22          & 3.98 & 6.55 & 4.58 \\
\midrule
\TRADE{}, $w\!=\!0.1$                       & 3.21          & 3.88          & 6.52          & 4.54 \\
\TRADE{}, $w\!=\!0.3$                            & 3.15          & \textbf{3.80}          & \textbf{6.52}          & \textbf{4.49} \\
\TRADE{}, $w\!=\!0.5$                            & \textbf{3.13}          & 3.84          & 6.60          & 4.52 \\
\TRADE{}, $w\!=\!0.7$                   & 3.15 & 3.86          & 6.64          & 4.55 \\
\TRADE{}, $w\!=\!0.9$                    & 3.17          & 3.89          & 6.67          & 4.58 \\
\midrule
Transducer ($w\!\to\!1$)         & 3.19          & 3.92          & 6.67          & 4.59 \\
\bottomrule
\end{tabular}%
}
\end{table}

\begin{table}[!htbp]
\small
\centering
\setlength{\tabcolsep}{6pt}
\caption{Vocabulary size ablation. 8-set Open ASR Leaderboard mean WER (\%) under three decoding modes. Bold = per-column minimum.}
\label{tab:vocab_fused}
\begin{tabular}{lccc}
\toprule
\textbf{Vocab} & \textbf{\TRADE{}} & \textbf{Decoder mode} & \textbf{Stream-640\,ms} \\
\midrule
10K & 6.94          & 7.09          & 10.78          \\
15K & 6.92          & 6.85          & 11.56          \\
20K & \textbf{6.71} & \textbf{6.43} & \textbf{8.97}  \\
\bottomrule
\end{tabular}
\end{table}
% ===============================================================
\subsection{End-of-Utterance Detection}
\label{sec:eou}

Since \TRADE{} emits tokens with acoustic timestamps, we exploit its streaming output for real-time end-of-utterance (EOU) detection.
We decode the 11 TED-LIUM test talks~\citep{hernandez2018tedlium3} as unsegmented long-form audio in 320\,ms streaming mode, yielding 1{,}094 reference boundaries, and score predictions with greedy 1-to-1 matching at tolerance $\tau\!=\!0.5$\,s.
We compare three predictors: \emph{VAD-only} (Silero VAD~\citep{silero2021vad} silence onsets); \emph{Punctuation-only} (terminal tokens $\in \{\text{``.''}, \text{``?''}, \text{``!''}\}$ from \TRADE{}); and \emph{Symmetric fusion} (proposed), which fires only when a terminal-or-weak punctuation token ($\in \{\text{``.''}, \text{``?''}, \text{``!''}, \text{'',''},  \text{``;''}, \text{``:''}\}$) and a Silero silence onset co-occur within window $\delta$, using VAD for timing and punctuation as a semantic gate to suppress spurious gaps.

%\paragraph{Results.}
%Table~\ref{tab:eou} summarises the best configuration per family at $\text{tol}=0.5$\,s.

For each family we report the best configuration found by grid search:
\emph{VAD-only} uses Silero with speech-probability threshold $0.5$ and minimum silence $30$\,ms;
\emph{Punctuation-only} uses the terminal set $\{.\,?\,!\}$;
\emph{Symmetric fusion} uses the extended set $\{.\,?\,!\,,\,;\,:\}$ together with Silero ($0.9/20$\,ms) and co-occurrence window $\delta\!=\!0.5$\,s.

\begin{table}[!htbp]
\centering
\small
\caption{End-of-utterance detection on TED-LIUM, \TRADE{} 320\,ms streaming decode. \textbf{P}: precision; \textbf{R}: recall. Best configuration per family shown (see text).}
\label{tab:eou}
\begin{tabular}{lccc}
\toprule
\textbf{Method} & \textbf{P} & \textbf{R} & \textbf{F\textsubscript{1}} \\
\midrule
VAD-only~\citep{silero2021vad}    & 0.336 & 0.684 & 0.451 \\
Punctuation-only                  & 0.216 & 0.793 & 0.340 \\
\midrule
\textbf{Symmetric fusion (ours)}  & \textbf{0.362} & \textbf{0.724} & \textbf{0.482} \\
\bottomrule
\end{tabular}
\end{table}

Symmetric fusion achieves $F_1 = 0.482$, outperforming both baselines by at least $+0.03$ absolute $F_1$, with p95 detection latency of 0.416\,s (consistent with the 320\,ms chunk stride plus one lookahead, total ${\approx}640$\,ms commit delay).

%\input{sections/discussion}

%\FloatBarrier
% ===============================================================
\section{Conclusion}
\label{sec:conclusion}

We presented \TRADE{}, a multimodal LLM augmented with a transducer branch that gives the system frame-synchronous acoustic alignment without sacrificing the LLM's linguistic reasoning.
The key design choices — dual tightly-coupled vocabularies, chunk-synchronized streaming training, and localized decoder audio attention — address the core obstacles to deploying a Speech LLM in real-time settings.
A single \TRADE{} checkpoint supports three distinct operating modes at inference time: LLM-only decoding, offline decoding, and streaming decoding across a continuous range of latency operating points, from 320\,ms to fully offline.

Experimentally, \TRADE{} matches or exceeds strong published baselines on the Open ASR Leaderboard, transcribes long-form audio natively without external segmentation, and enables real-time utterance boundary detection by fusing the transducer's punctuation emissions with acoustic voice-activity signals.

We believe the transducer–LLM coupling paradigm demonstrated in \TRADE{} is a promising direction for building Speech LLMs that are simultaneously accurate, streamable, and long-form capable — properties that have previously required separate, specialized models.

% ===============================================================
\section{Limitations}
\label{sec:limitations}

\paragraph{English-only evaluation.}
All experiments are conducted on English speech. The transducer vocabulary derivation procedure — pruning non-verbalized tokens and merging pronunciation-equivalent surface forms — is language-specific, and extending \TRADE{} to other languages requires rebuilding the compact vocabulary and retraining the joint network. Multilingual capability has not been evaluated.

\paragraph{Streaming accuracy degradation.}
While \TRADE{} supports streaming across a range of chunk sizes, there is a meaningful WER gap between offline and low-latency streaming operation (e.g., 6.71\% vs.\ 9.35\% average WER on the Open ASR Leaderboard at 640\,ms chunks). For latency-sensitive applications, users must accept this accuracy trade-off.

%\paragraph{Long-form context fragility.}
%Long-form decoding performance is sensitive to the \texttt{max\_compute\_token\_length} hyperparameter: reducing the LLM's text context window from 20 to 50 tokens causes WER to degrade 3--5$\times$ across all models and datasets. This fragility indicates that the LLM's conditioning on prior decoded text is not yet robust, and tuning this parameter for new domains or recording conditions requires careful validation.

\paragraph{End-of-utterance detection.}
The EOU detection experiments are conducted on a single dataset (TED-LIUM, 11 talks) with ground truth derived from an automatic segmenter rather than human-labeled utterance boundaries. The achieved $F_1 = 0.48$ is moderate, and performance on conversational or spontaneous speech — where utterance boundaries are less acoustically distinct — is not evaluated.

\paragraph{Compute requirements.}
Training \TRADE{} requires 16$\times$H200 GPUs and thirty-five thousand optimizer steps on large-scale data (${\approx}153$K hours). This scale of compute may limit reproducibility for researchers without access to equivalent hardware. We report a single checkpoint per configuration without statistical significance tests or error bars across seeds.

\paragraph{Scope of comparison.}
The Open ASR Leaderboard comparison includes models of varying sizes and training data scales. \TRADE{} is built on a 1B-parameter LLM backbone and has not been scaled to larger LLMs; the benefits of the transducer–LLM coupling at larger scales remain to be validated.

%\input{latex/sections/broader_impacts}

\bibliography{dart_paper}

\clearpage
\appendix

% ===============================================================
\section{Detailed Model Configuration}
\label{app:config}

\paragraph{Encoder.}
The acoustic backbone is a FastConformer-XL encoder initialized from Parakeet-TDT-0.6B-v2~\citep{shah2025canary} (24 layers, 1{,}024-dim hidden, 8$\times$ subsampling).
The top six transformer layers are fine-tuned; the remaining layers are frozen.
SpecAugment is applied during training.

\paragraph{Encoder-decoder adaptor.}
A single-layer causal transformer with 2$\times$ frame-stacking downsampling projects encoder outputs into the LLM's embedding space, making the LLM path fully streaming-compatible.

\paragraph{LLM.}
The LLM backbone is Llama-3.2-1B~\citep{grattafiori2024llama}, fine-tuned with LoRA~\citep{hu2021lora} ($r=16$, $\alpha=32$) on all attention projections.

\paragraph{Encoder-to-joint adaptor.}
A single linear projection maps encoder frame embeddings into the joint network's input space.

\paragraph{Decoder-to-joint adaptor.}
A single linear projection maps the LLM's last hidden states (at positions used to predict verbalized tokens) into the joint network's prediction space.

\paragraph{Joint network.}
A single-hidden-layer multi-layer perceptron (MLP) with ReLU activation combines the encoder and decoder projections.
The joint dimension is 1{,}024.
All adaptor and joint-network parameters are trained from scratch.
The transducer operates over a compact 20K-token vocabulary derived from the LLM vocabulary via pronunciation normalization (see Appendix~\ref{app:vocab_construction}).

\paragraph{Input preparation.}
The LLM input sequence is structured as:
\begin{center}
\texttt{[optional context]} \texttt{Transcribe the speech.} \texttt{<|audioplaceholder|>}
\end{center}
where \texttt{<|audioplaceholder|>} is a special locator token whose embedding is replaced by the LLM audio embeddings at that position.
In Phase~1 (plain ASR), no context is prepended and the prompt is fixed to \texttt{"Transcribe the speech."}.
In Phase~2 (contextual ASR), the transcription of the preceding utterance is prepended as context with a cutoff length of 200 tokens, applied with 50\% probability per sample.
Audio is pre-segmented; individual utterances are capped at 30\,s, batches are packed up to 240\,s total duration, and each sample is limited to at most 80 samples per mini-batch.

\paragraph{Optimizer.}
All models are trained with AdamW ($\beta_1=0.9$, $\beta_2=0.98$, weight decay $10^{-3}$) under a cosine annealing schedule with a 500-step linear warmup, minimum lr $10^{-6}$, and $35{,}000$ total steps.
Per-group learning rates: the unfrozen encoder top-6 layers and LoRA parameters use a multiplier of $0.1$ (effective lr $10^{-4}$); all adaptor and joint-network parameters use a multiplier of $1.0$ (effective lr $10^{-3}$).
Gradients are clipped to $1.0$ and accumulated over 8 steps.
Training uses 16$\times$H200 GPUs with bfloat16 mixed precision.

\paragraph{Chunk-size training.}
Dynamic chunk-size training is applied throughout.
For utterances up to 25\,s, the chunk size is sampled each step from the discrete set $\{4,\,8,\,16,\,24,\,32,\,\text{full}\}$ (post-subsampling frames) with probabilities $\{0.1,\,0.1,\,0.1,\,0.1,\,0.1,\,0.5\}$; the full-context option (50\%) reduces to standard offline training.
Utterances longer than 25\,s are always trained with the full context.
This multi-granularity exposure provides robustness across latency operating points.

% ===============================================================
\section{Transducer Vocabulary Construction}
\label{app:vocab_construction}

The compact transducer vocabulary $\mathcal{V}_\text{trans}$ is derived from the LLM vocabulary $\mathcal{V}_\text{llm}$ ($|\mathcal{V}_\text{llm}|\!=\!128{,}000$ for Llama-3.2-1B-Instruct, with 280{,}147 BPE merge rules) in three stages: corpus tokenization, token frequency estimation, and frequency-guided pruning.

\paragraph{Stage 1: Corpus text sampling and tokenization.}
A random 50\% sample of the training-set transcripts is drawn, with any line containing written-form numeric content (digits, currency symbols, time expressions, years, percentages, ordinals) discarded so that the frequency statistics reflect spoken-style text only.
The retained transcripts are tokenized with the LLM tokenizer to produce sequences of token surface forms.
This places the frequency statistics in the same BPE token space as the LLM, including the space-prefix marker (the Ġ\,/\,U+0120 character prepended to word-initial tokens in Llama's BPE scheme).

\paragraph{Stage 2: Token normalization and frequency estimation.}
Before counting, each token surface form is normalized: the BPE space prefix is stripped, the token is lowercased, and leading/trailing punctuation (except apostrophes) is removed.
Tokens that reduce to the empty string after normalization — \ie{} those containing no alphanumeric character — are classified as \emph{non-verbalized} and handled separately.
The remaining normalized forms are counted across the corpus, yielding a ranked frequency list over acoustically realizable word-piece types.
Normalizing before counting ensures that surface variants that share the same spoken realization (\eg{} ``\textit{Hello}'' and ``\textit{hello}'', or ``\textit{Ġworld}'' and ``\textit{world}'') are counted together rather than as separate entries.
On our training corpus this procedure yields \textbf{29{,}578 unique normalized token types}, an upper bound on $|\mathcal{V}_\text{trans}|$ achievable without dropping any observed surface form.

\paragraph{Stage 3: Frequency-guided pruning.}
The top-$K$ most frequent normalized forms are kept.
In our experiments $K\!=\!20{,}000$, giving $|\mathcal{V}_\text{trans}| \approx 20$K.

\emph{Non-verbalized tokens (always kept).}
Tokens whose normalized form contains no alphanumeric character --- pure whitespace, newlines, punctuation-only sequences, and formatting symbols --- are unconditionally retained in the pruned tokenizer regardless of frequency.
Removing them would corrupt the LLM tokenizer's ability to reconstruct arbitrary text.
They do not appear in the transducer output or the joint lattice; they are assigned an empty entry in the mapping table (see below) and excluded from the transducer's verbalized vocabulary by design.

\paragraph{Tokenizer construction.}
The LLM tokenizer uses Byte-Pair Encoding~\citep{sennrich2016bpe}, where longer tokens are produced by iteratively merging shorter constituents.
To produce the compact tokenizer, we keep the original vocabulary in place but mark pruned entries as inactive; tokenization of new text greedily selects only the longest \emph{active} vocabulary entry at each position, so no inactive form can ever appear in a transducer-side output.
Merge rules that reference an inactive token (either as an input or as the merged output) are dropped for consistency, while inactive entries themselves are retained at their original positions so that no surviving token's identifier is shifted.
Crucially, \emph{all original LLM token IDs are preserved}: pruning never reassigns identifiers, so the LLM's embedding table and output projection remain valid without remapping, and every token shared between $\mathcal{V}_\text{trans}$ and $\mathcal{V}_\text{llm}$ keeps the same ID in both vocabularies.

\paragraph{Mapping table.}
The final output of the pruning step is a two-column mapping table over $\mathcal{V}_\text{llm}$: each LLM token maps to its normalized transducer form, or to an empty entry if it is disabled or non-verbalized.
This table is loaded at training and inference time to project transducer posteriors into the LLM vocabulary for the joint scoring step described in Section~\ref{sec:formatted}.

\paragraph{Empirical statistics across vocabulary sizes.}
Table~\ref{tab:vocab_stats} reports the result of running the pipeline at six target sizes $K \in \{5{,}000, 10{,}000, 15{,}000, 20{,}000, 25{,}000, 30{,}000\}$ on the same corpus, evaluated on a held-out 100K-transcription set.
At each $K$, \emph{Vocab Size} $|\mathcal{V}_\text{trans}|$ is the number of unique active tokens in the pruned tokenizer (slightly larger than $K$ due to the short-token and non-verbalized tokens always retained); \emph{Mapped Vocab} is the number of LLM token slots that survive pruning; \emph{BPE Rules} is the number of merge rules in the resulting tokenizer (vs.\ 280{,}147 for the base LLM); and \emph{Avg Tok/Sample} is the mean token count per transcription on the 100K-sample (vs.\ 28.8 for the base LLM tokenizer on the same set).

\begin{table*}[t]
\small
\centering
\caption{Compact tokenizers produced from the Llama-3.2-1B-Instruct vocabulary, evaluated on a 100K-transcription sample. Vocab size, mapped vocab, and BPE rules scale roughly linearly with $K$; avg tok/sample decays quickly toward the LLM baseline of 28.8 — by $K\!=\!15{,}000$ inflation is already under 4\%, and beyond $K\!=\!25{,}000$ it is essentially indistinguishable from the LLM.}
\label{tab:vocab_stats}
\begin{tabular}{lrrrr}
\toprule
\textbf{Tokenizer} & \textbf{Vocab Size} & \textbf{Mapped Vocab} & \textbf{BPE Rules} & \textbf{Avg Tok/Sample} \\
\midrule
LLM (Llama-3) &  128{,}000 &       — & 280{,}147 & 28.8 \\
\midrule
\texttt{K=5000}  &   5{,}087 & 25{,}681 &  50{,}236 & 34.0 \\
\texttt{K=10000} &  10{,}084 & 38{,}006 &  80{,}524 & 30.8 \\
\texttt{K=15000} &  15{,}080 & 47{,}216 & 105{,}322 & 29.8 \\
\texttt{K=20000} &  20{,}077 & 54{,}920 & 127{,}553 & 29.3 \\
\texttt{K=25000} &  25{,}077 & 62{,}531 & 151{,}172 & 29.0 \\
\texttt{K=30000} &  29{,}655 & 69{,}866 & 175{,}243 & 28.9 \\
\bottomrule
\end{tabular}
\end{table*}

The token-count inflation is not uniform across word types.
Common function words and high-frequency content words that survive the top-$K$ cutoff produce identical token sequences to the LLM tokenizer; inflation concentrates in mid-to-low-frequency multi-syllable words whose intermediate BPE merges have been pruned.
For example, at $K\!=\!10{,}000$, \texttt{infrastructure} tokenizes as \texttt{inf|ra|str|u|ct|ure} (6 tokens) instead of the LLM's \texttt{inf|rastructure} (2), and \texttt{telecommunications} as \texttt{tele|comm|un|ic|ations} (5) instead of \texttt{tele|communications} (2).
This explains why average tokens-per-sample drops sharply between $K\!=\!5{,}000$ and $K\!=\!10{,}000$ (34.0 \textrightarrow{} 30.8) and continues to fall thereafter, but with diminishing returns: by $K\!=\!15{,}000$ inflation is already under 4\%, and the marginal long words added at $K\!=\!20{,}000$ and beyond contribute progressively less to the average.

We adopt $K\!=\!20{,}000$ as the default for \TRADE{}.
The fused-decoding vocabulary size ablation in Section~\ref{sec:ablation} (Table~\ref{tab:vocab_fused}) and the decoder-/streaming-mode comparison in Appendix~\ref{app:vocab_ablation} both confirm this choice: across all three decoding modes, $K\!=\!20{,}000$ matches or improves on smaller vocabularies, with the largest gains under streaming where the larger vocabulary captures more rare-word evidence under bounded chunk context.
At this size, the resulting joint-network output ($\sim$20K classes) remains tractable, while the BPE fragmentation of mid-frequency words is essentially eliminated (Table~\ref{tab:vocab_stats}).

% ===============================================================
\section{Chunk-aware Encoder: CADA Details}
\label{app:cada}

The Copy-and-Append Data Augmentation (CADA) scheme~\citep{liu2021crossattention, tang2025chunk} enables the Conformer encoder to incorporate exactly one chunk of lookahead context per encoder layer without cascading future information across layers or violating streaming causality.

\paragraph{Input augmentation.}
Given an $N$-frame input sequence and chunk size $C$, the encoder internally constructs an augmented sequence of length $2N{-}C$ by appending $N{-}C$ \emph{copy frames} — duplicates of frames $[C, N)$ — to the original sequence. The copy frames represent the lookahead content of the next chunk.

\paragraph{CADA attention mask.}
A block-diagonal boolean attention mask governs which frames may attend to which. The visibility rules are:
\begin{itemize}
  \item \emph{Original $\to$ original}: frame $i$ attends causally to all preceding original frames (same chunk and earlier).
  \item \emph{Original $\to$ copy}: frame $i$ in chunk $\lfloor i/C \rfloor$ may attend to the lookahead copy of the immediately following chunk only; it cannot see copies of any further chunk.
  \item \emph{Copy $\to$ original}: copy frame $j$ (true temporal position $j{+}C$, assigned to copy-chunk $\lfloor (j{+}C)/C \rfloor$) may attend to all original frames up to and including its assigned chunk.
  \item \emph{Copy $\to$ copy}: copy frame $j$ attends only within its own copy-chunk, causally.
\end{itemize}
This ensures exactly one chunk of lookahead is exposed per layer, with no compounding across layers.

\paragraph{Relative positional encodings.}
Standard relative positional encodings assume consecutive temporal positions, which the copy frames violate. CADA replaces the default relative-shift computation with explicit indexing: each augmented frame is assigned its true temporal position ($i$ for originals, $i{+}C$ for copies), and relative distances $\Delta_{i,j} = t_i - t_j$ are computed directly from these true positions.

\paragraph{Per-chunk convolution.}
Conformer convolution modules are applied per chunk with explicit left-context and lookahead padding rather than over the full augmented sequence, maintaining identical causal constraints as the self-attention masking.

\paragraph{Outputs.}
After all encoder layers, the augmented sequence is split back into the \textbf{original encoding} $\mathbf{H}_\mathrm{orig} \in \mathbb{R}^{B \times D \times T_\mathrm{enc}}$ and the \textbf{lookahead encoding} $\mathbf{H}_\mathrm{look} \in \mathbb{R}^{B \times D \times (T_\mathrm{enc}-C)}$, where copy chunk $k$ in $\mathbf{H}_\mathrm{look}$ represents the encoder's anticipatory view of original chunk $k{+}1$. No new learnable parameters are introduced; all weights are shared with the base encoder.

\paragraph{CADA-aware adaptor.}
The encoder-decoder adaptor processes $\mathbf{H}_\mathrm{orig}$ and $\mathbf{H}_\mathrm{look}$ jointly by concatenating them along the time axis and applying causal transformer layers with a dedicated block-diagonal mask. The mask differs from the encoder's in one key aspect: original frames are \emph{blocked} from attending to any copy frames, keeping adapted originals strictly causal. Copy chunk $k$ attends to all original chunks $0,\ldots,k$ and to copy frames within its own chunk only. After processing, the output is split back into adapted originals and adapted copies; the LLM receives the adapted copies as an anticipatory prefix, gaining one chunk of lookahead context without any future-frame leakage.

% ===============================================================
\section{Evaluation and Training Data}
\label{app:data}

\subsection{Long-Form Evaluation Datasets}
\label{app:longform_data}

Table~\ref{tab:longform_data} summarises the three long-form evaluation benchmarks used in Section~\ref{sec:longform}.
Each dataset consists of full-length recordings with durations ranging from roughly 7 minutes to over 2 hours, posing a substantially different challenge from the short-form utterances used in standard ASR benchmarks.
Earnings-21~\citep{delrio2021earnings21} and Earnings-22~\citep{delrio2022earnings22} are collections of earnings call recordings covering spontaneous, domain-specific financial speech with frequent domain terminology, cross-talk, and variable audio quality.
TED-LIUM~3~\citep{hernandez2018tedlium3} consists of TED conference talks — relatively clean, prepared speech — used here in its test split of 11 talks.

\begin{table*}[t]
\centering
\small
\caption{Long-form audio evaluation datasets.}
\label{tab:longform_data}
\begin{tabular}{lrrrr}
\toprule
\textbf{Dataset} & \textbf{Recordings} & \textbf{Min (min)} & \textbf{Max (min)} & \textbf{Mean (min)} \\
\midrule
Earnings-21~\citep{delrio2021earnings21}  &  44 & 18.29 &  95.68 & 53.54 \\
Earnings-22~\citep{delrio2022earnings22}  & 125 & 14.58 & 123.45 & 57.55 \\
TED-LIUM~3~\citep{hernandez2018tedlium3}  &  11 &  6.89 &  29.53 & 16.74 \\
\bottomrule
\end{tabular}
\end{table*}

\subsection{Large-Scale Training Data}
\label{app:train_data}

Table~\ref{tab:data} summarises the composition of the large-scale multi-domain corpus used for the second \TRADE{} model (Section~\ref{sec:experiments}).

\begin{table*}[t]
\centering
\small
\caption{Large-scale multi-domain training corpus composition. Hours and percentages are computed from audio durations in the dataset manifest.}
\label{tab:data}
\begin{tabular}{llrr}
\toprule
\textbf{Source} & \textbf{Domain} & \textbf{Hours} & \textbf{\%} \\
\midrule
Granary YODAS English portion~\citep{koluguri2025granary} & Web video / diverse  & 102{,}461 & 66.8 \\
Multilingual LibriSpeech EN~\citep{pratap2020mls}             & Audiobooks           &  44{,}420 & 29.0 \\
SPGISpeech~\citep{oneill2021spgi}                             & Financial calls      &   4{,}818 &  3.1 \\
LibriSpeech~\citep{panayotov2015librispeech} (train-other)    & Audiobooks           &     497   &  0.3 \\
VoxPopuli EN~\citep{wang2021voxpopuli}                        & Parliamentary speech &     462   &  0.3 \\
LibriSpeech (train-clean-360)                                 & Audiobooks           &     364   &  0.2 \\
Earnings-22~\citep{delrio2022earnings22}                      & Earnings calls       &     115   &  0.1 \\
LibriSpeech (train-clean-100)                                 & Audiobooks           &     101   &  0.1 \\
AMI~\citep{mccowan2005ami} (IHM)                              & Meeting room         &      87   &  0.1 \\
AMI (SDM)                                                     & Meeting room         &      86   &  0.1 \\
\midrule
\textbf{Total}                                                &                      & \textbf{153{,}411} & \textbf{100.0} \\
\bottomrule
\end{tabular}
\end{table*}

% ===============================================================
\section{Long-Form Decoding Configuration}
\label{app:longform_decoding}

This section details the inference recipe used for the long-form ASR results in Section~\ref{sec:longform}.
Three additions on top of the short-form streaming recipe bound inference-time memory and prevent runaway emission on hours-long inputs: chunked acoustic streaming, an LLM token sliding window, and a runtime repetition filter.

\paragraph{Chunked acoustic streaming.}
The acoustic encoder consumes the input chunk-by-chunk via the CADA encoder's incremental forward (Appendix~\ref{app:cada}), carrying per-layer self-attention KV state forward across chunks.
At each chunk boundary, the LLM is re-prefilled on the prompt, the windowed audio embeddings $h_{\tau^-_{\delta(t)}{:}\tau^+_{\delta(t)}}$ (LDAA, Section~\ref{sec:windowed}), and the active transcript token buffer.
We use $5{,}120$\,ms chunks with a $64$-frame ($\approx 5.12$\,s) left-context window matching LDAA.
This bounds both encoder- and decoder-side memory independently of input length.

\paragraph{LLM token sliding window.}
The LLM prefill input is capped at the $20$ most recently decoded tokens at each chunk boundary.
Older tokens are shifted to a side buffer and re-concatenated into the final transcript, but no longer participate in LLM conditioning at subsequent chunks.
Without this cap, the LLM accumulates tens of thousands of decoded tokens over a one-hour call, and the prefill quickly becomes dominated by stale early context that triggers repetition and hallucination.

\paragraph{Repetition filter.}
After every emission, an $n$-gram-loop detector scans the active sliding window for suffix cycles up to length~3, requiring at least $8$ tokens of total cycle span and $3$ repetitions before firing.
On a hit, the detected suffix cycle is dropped from the active buffer, the LLM KV cache is cropped to the pre-cycle prefix, and the decoder is forced to advance one acoustic frame.
Tokens that have already shifted out of the active window into the side buffer are never modified.

% ===============================================================
\section{Emission Timing Analysis}
\label{app:emission}

To validate the localized audio attention window and characterize how \TRADE{} aligns token emission with acoustic evidence, we analyse the trained model on LibriSpeech \texttt{test\_other}, evaluated at three streaming chunk sizes: 320, 640, and 5{,}120\,ms. Word-level reference alignments are obtained from Parakeet-CTC-1.1B via NeMo Forced Aligner (NFA, 80\,ms output stride).

\paragraph{Acoustic support span.}
LDAA (Section~\ref{sec:windowed}) bounds the LLM's audio context to a fixed-duration sliding window; the window must be wide enough to cover the acoustic evidence each token requires.
We quantify this requirement by computing the \emph{acoustic support span} of each word: the interval from the word's onset (per NFA reference alignment) to the transducer's emission timestamp for that word.
This span measures how far back into the audio the model effectively looked before committing to the emission, and its distribution directly determines how large the LDAA window must be.
Table~\ref{tab:hull} reports support-span statistics for the large-scale multi-domain model.

\begin{table}[t]
\centering
\small
\caption{Acoustic support-span statistics on LibriSpeech \texttt{test\_other}. Values are reported for the 320\,ms streaming mode and are invariant to chunk size across 320, 640, and 5{,}120\,ms.}
\label{tab:hull}
\begin{tabular}{cccc}
\toprule
\textbf{Mean} & \textbf{95th pct.} & \textbf{99th pct.} & \textbf{Max} \\
\midrule
1.54\,s & 2.56\,s & 3.28\,s & 14.80\,s \\
\bottomrule
\end{tabular}
\end{table}

The model achieves a mean support span of 1.54\,s and a 95th-percentile of 2.56\,s. Crucially, \textbf{a 3.5\,s window covers the 99th-percentile span}, and the 5\,s default provides a comfortable margin. Support-span statistics are \textbf{invariant to chunk size} across all three streaming configurations, confirming that the acoustic context requirement is independent of the streaming latency operating point.

\paragraph{End-of-utterance latency.}
We introduce an end-of-utterance (EOU) latency metric $\Delta_\text{last} = t_\text{last emit} - t_\text{last NFA word end}$, where positive values indicate the transducer trails the alignment reference.
At 320\,ms streaming, 95.4\% of utterances emit within $+200$\,ms of the reference end (p5/p95: $[-320, +160]$\,ms), confirming near-real-time end-pointing with no systematic trailing delay. The 95th-percentile shifts by only ${\sim}$80\,ms from 320\,ms to 5{,}120\,ms chunks — one encoder frame of additional look-ahead — further demonstrating that timing precision is largely insensitive to chunk size.

Together, these results confirm that the 5\,s localized attention window is well-calibrated to the model's actual acoustic context requirements, and that end-pointing latency is near-zero in the median case across all streaming operating points.

% ===============================================================
\section{Vocabulary Size Ablation: Per-Testset Breakdown}
\label{app:vocab_ablation}

Section~\ref{sec:ablation} reports the 8-set mean WER for the vocabulary size ablation across three decoding modes.
Table~\ref{tab:ablation} gives the per-testset numbers underlying that summary, using the same checkpoints and training setup of Section~\ref{sec:experiments}.

\begin{table*}[t]
\small
\centering
\setlength{\tabcolsep}{4.5pt}
\caption{Vocabulary size ablation, per-testset WER (\%) on the Open ASR Leaderboard English benchmark. $\dagger$~mean over the eight sets. Bold = per-mode minimum in each column.}
\label{tab:ablation}
\begin{tabular}{lrcccccccc}
\toprule
\textbf{Vocab} & \textbf{Mean}$^\dagger$ & \textbf{AMI} & \textbf{E22} & \textbf{Giga} & \textbf{LS-c} & \textbf{LS-o} & \textbf{SPGI} & \textbf{TED} & \textbf{Vox} \\
\midrule
\multicolumn{10}{l}{\emph{\TRADE{}}} \\
\midrule
10K & 6.94          & 16.13          & 11.31          & 10.29 & \textbf{1.57} & 3.13          & 2.39          & \textbf{3.61} & 7.09          \\
15K & 6.92          & 16.11          & \textbf{10.96}          & 10.44          & 1.59          & \textbf{2.99} & 2.48          & 3.93          & 6.68          \\
20K & \textbf{6.71} & \textbf{14.85} & 11.02 & \textbf{10.24}          & 1.60 & 3.13          & \textbf{2.36} & 3.84          & \textbf{6.60} \\
\midrule
\multicolumn{10}{l}{\emph{Decoder mode}} \\
\midrule
10K & 7.09          & 17.47          & 11.26          & 10.08 & 1.59          & 3.10          & 2.20          & \textbf{3.50} & 7.09          \\
15K & 6.85          & 16.58          & 10.92 & 10.23          & 1.60          & \textbf{3.02} & 2.24          & 3.67          & 6.55          \\
20K & \textbf{6.43} & \textbf{13.67} & \textbf{10.88}          & \textbf{9.85}          & \textbf{1.68} & 3.26          & \textbf{2.14} & 3.64          & \textbf{6.34} \\
\midrule
\multicolumn{10}{l}{\emph{Streaming (640\,ms chunk)}} \\
\midrule
10K & 10.78          & 22.99          & 18.58          & 12.63          & 2.53          & 5.36          & 6.26          & 6.86          & 10.99          \\
15K & 11.56          & 24.87          & 19.29          & 13.31          & 2.33 & 5.05 & 7.08          & 8.08          & 12.50          \\
20K & \textbf{8.97}  & \textbf{18.04} & \textbf{16.23} & \textbf{11.25} & \textbf{2.29}          & \textbf{5.00}          & \textbf{4.60} & \textbf{4.98} & \textbf{9.35}  \\
\bottomrule
\end{tabular}
\end{table*}

Across all three decoding modes, $|\mathcal{V}_\text{trans}|\!=\!20{,}000$ delivers the best 8-set mean.
Under \TRADE{} decoding, 20K wins four of the eight per-set columns; the largest single-set gain is on AMI ($-1.28$ vs 10K).
Decoder mode shows a similar pattern (20K wins six per-set columns with the largest gain again on AMI, $-3.80$ vs 10K).
Streaming exposes the gap most dramatically: 20K reaches $8.97\,\%$ mean WER, $-1.81$ abs vs 10K, and wins every single per-set column.
%The non-monotone behavior at 15K under streaming ($+0.78$ vs 10K) is driven by long-form-style sets (AMI, Earnings-22, VoxPopuli), where the 20K vocabulary captures more rare-word evidence under bounded chunk context while 15K fragments mid-frequency words just enough to amplify chunk-related context loss.

% ===============================================================
\section{Streaming Latency Analysis}
\label{app:latency}

We evaluate streaming latency for \TRADE{} on LibriSpeech \texttt{dev-other} across six chunk sizes from 320\,ms to 5{,}120\,ms, using the SimulEval toolkit~\citep{ma2020simuleval}.
The decode recipe uses joint streaming decoding with fusion weight $w\!=\!0.5$, blank penalty $0.5$, and a 64-frame left-context window (${\approx}5.12$\,s).
We report three complementary latency metrics beyond WER--AL (Figure~\ref{fig:wer_al} in Section~\ref{sec:leaderboard}):
\textbf{Differentiable AL (DAL)}~\citep{cherry2019thinking}, a monotone-enforced variant of AL used as a training proxy;
\textbf{Average Proportion (AP)}~\citep{cho2016simultaneous}, a unit-free $[0,1]$ score where 1.0 means the model waits for the full audio before each emission (fully offline); and
\textbf{Real-Time Factor (RTF)}, the wall-clock decoding time divided by audio duration on a single H200 GPU.

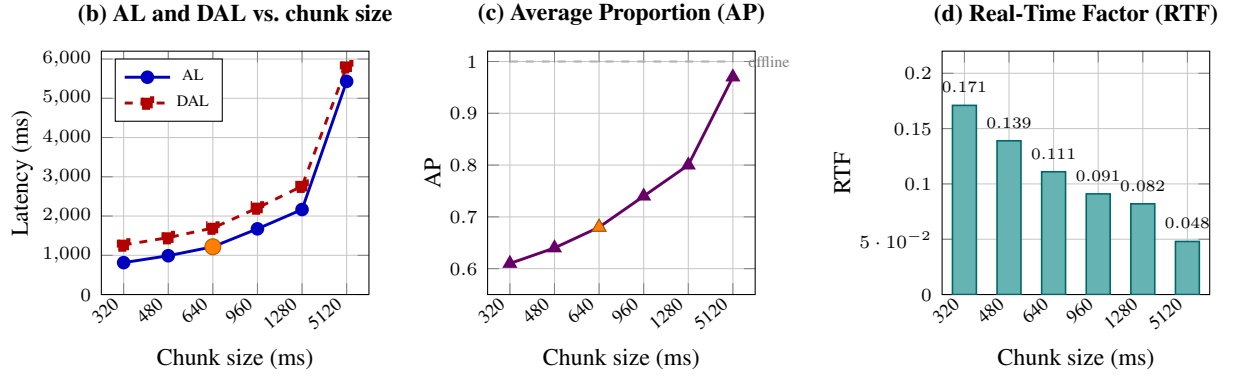
\begin{figure*}[t]
\centering
% --- Three panels side by side ---
\begin{minipage}[t]{0.32\textwidth}
\centering
\begin{tikzpicture}
\begin{axis}[
    width=\linewidth,
    height=4.8cm,
    xlabel={Chunk size (ms)},
    ylabel={Latency (ms)},
    symbolic x coords={320,480,640,960,1280,5120},
    xtick=data,
    xticklabel style={rotate=40, anchor=east, font=\scriptsize},
    ymin=0, ymax=6200,
    ytick={0,1000,2000,3000,4000,5000,6000},
    yticklabel style={font=\scriptsize},
    grid=both,
    grid style={line width=0.2pt, draw=gray!25},
    major grid style={line width=0.35pt, draw=gray!45},
    label style={font=\small},
    xlabel style={yshift=2pt},
    ylabel style={yshift=-4pt},
    legend style={font=\tiny, at={(0.05,0.97)}, anchor=north west},
    clip=false,
    title style={font=\small\bfseries},
    title={(b) AL and DAL vs.\ chunk size},
]
\addplot[
    color=blue!75!black,
    mark=*,
    mark size=2pt,
    line width=1.1pt,
] coordinates {
    (320, 813)
    (480, 989)
    (640, 1217)
    (960, 1674)
    (1280, 2166)
    (5120, 5431)
};
\addlegendentry{AL};
\addplot[
    color=red!70!black,
    mark=square*,
    mark size=2pt,
    line width=1.1pt,
    dashed,
] coordinates {
    (320, 1263)
    (480, 1451)
    (640, 1694)
    (960, 2206)
    (1280, 2761)
    (5120, 5804)
};
\addlegendentry{DAL};
% highlight 640ms
\addplot[
    color=orange!90!black,
    mark=*,
    mark size=3pt,
    only marks,
    mark options={fill=orange, draw=orange!70!black},
] coordinates {(640, 1217)};
\end{axis}
\end{tikzpicture}
\end{minipage}\hfill
\begin{minipage}[t]{0.32\textwidth}
\centering
\begin{tikzpicture}
\begin{axis}[
    width=\linewidth,
    height=4.8cm,
    xlabel={Chunk size (ms)},
    ylabel={AP},
    symbolic x coords={320,480,640,960,1280,5120},
    xtick=data,
    xticklabel style={rotate=40, anchor=east, font=\scriptsize},
    ymin=0.55, ymax=1.02,
    ytick={0.6,0.7,0.8,0.9,1.0},
    yticklabel style={font=\scriptsize},
    grid=both,
    grid style={line width=0.2pt, draw=gray!25},
    major grid style={line width=0.35pt, draw=gray!45},
    label style={font=\small},
    xlabel style={yshift=2pt},
    ylabel style={yshift=-4pt},
    clip=false,
    title style={font=\small\bfseries},
    title={(c) Average Proportion (AP)},
]
\addplot[
    color=violet!80!black,
    mark=triangle*,
    mark size=2.2pt,
    line width=1.1pt,
] coordinates {
    (320,  0.61)
    (480,  0.64)
    (640,  0.68)
    (960,  0.74)
    (1280, 0.80)
    (5120, 0.97)
};
% reference line at 1.0 (offline)
\addplot[
    color=gray!60,
    dashed,
    line width=0.8pt,
    domain=320:5120,
] coordinates {(320,1.0)(5120,1.0)};
% highlight 640ms
\addplot[
    color=orange!90!black,
    mark=triangle*,
    mark size=3pt,
    only marks,
    mark options={fill=orange, draw=orange!70!black},
] coordinates {(640, 0.68)};
\node[font=\tiny, right=2pt, gray]  at (axis cs:5120, 1.0) {offline};
\end{axis}
\end{tikzpicture}
\end{minipage}\hfill
\begin{minipage}[t]{0.32\textwidth}
\centering
\begin{tikzpicture}
\begin{axis}[
    width=\linewidth,
    height=4.8cm,
    xlabel={Chunk size (ms)},
    ylabel={RTF},
    symbolic x coords={320,480,640,960,1280,5120},
    xtick=data,
    xticklabel style={rotate=40, anchor=east, font=\scriptsize},
    ymin=0, ymax=0.22,
    ytick={0,0.05,0.10,0.15,0.20},
    yticklabel style={font=\scriptsize},
    grid=both,
    grid style={line width=0.2pt, draw=gray!25},
    major grid style={line width=0.35pt, draw=gray!45},
    label style={font=\small},
    xlabel style={yshift=2pt},
    ylabel style={yshift=-4pt},
    bar width=9pt,
    nodes near coords,
    nodes near coords style={font=\tiny, yshift=1pt},
    every node near coord/.append style={/pgf/number format/fixed, /pgf/number format/precision=3},
    title style={font=\small\bfseries},
    title={(d) Real-Time Factor (RTF)},
]
\addplot[
    ybar,
    fill=teal!60,
    draw=teal!80!black,
    line width=0.6pt,
] coordinates {
    (320,  0.171)
    (480,  0.139)
    (640,  0.111)
    (960,  0.091)
    (1280, 0.082)
    (5120, 0.048)
};
\end{axis}
\end{tikzpicture}
\end{minipage}
\caption{Streaming latency metrics on LibriSpeech \texttt{dev-other}. Orange points mark the recommended 640\,ms operating point.
\textbf{(b)}~AL and DAL vs.\ chunk size; DAL is always $\geq$\,AL by construction since it enforces monotone emission.
\textbf{(c)}~Average Proportion (AP); the dashed line at 1.0 is the fully-offline reference.
\textbf{(d)}~RTF; all values are well below 1.0, confirming faster-than-realtime decoding on a single H200 across the full latency range.}
\label{fig:latency}
\end{figure*}

Table~\ref{tab:latency_numbers} provides the raw numbers for all four metrics across the six chunk sizes.
Figure~\ref{fig:latency} shows panels (b)--(d).

\begin{table}[t]
\centering
\small
\setlength{\tabcolsep}{4pt}
\caption{Streaming latency metrics across chunk sizes on LibriSpeech \texttt{dev-other}.}
\label{tab:latency_numbers}
\resizebox{\columnwidth}{!}{%
\begin{tabular}{rrrrrr}
\toprule
\textbf{Chunk (ms)} & \textbf{WER (\%)} & \textbf{AL (ms)} & \textbf{DAL (ms)} & \textbf{AP} & \textbf{RTF} \\
\midrule
 320          & 6.26          &  813          & 1263          & 0.61          & 0.171 \\
 480          & 4.83          &  989          & 1451          & 0.64          & 0.139 \\
 \textbf{640} & \textbf{4.04} & \textbf{1217} & \textbf{1694} & \textbf{0.68} & \textbf{0.111} \\
 960          & 3.60          & 1674          & 2206          & 0.74          & 0.091 \\
1280          & 3.56          & 2166          & 2761          & 0.80          & 0.082 \\
5120          & 2.95          & 5431          & 5804          & 0.97          & 0.048 \\
\bottomrule
\end{tabular}%
}
\end{table}

\paragraph{Trade-off shape.}
The 320\,ms to 480\,ms step yields the steepest WER reduction --- $1.43$\% absolute WER for an extra $176$\,ms of AL --- making low-latency operation surprisingly inexpensive; the 480\,ms to 640\,ms step delivers an additional $0.79$\% absolute for $228$\,ms more AL.
Past 640\,ms the curve flattens sharply: 960\,ms shaves an additional $0.44$\% absolute at the cost of $+457$\,ms of AL, and $1{,}280$\,ms is essentially indistinguishable from 960\,ms ($-0.04$\% absolute WER).
The recommended streaming operating point is \textbf{640\,ms} (AL\,=\,1{,}217\,ms, WER\,=\,4.04\,\%), which sits at the knee of the curve.
For latency-tolerant applications such as broadcast captioning or voicemail transcription, the $5{,}120$\,ms chunk (AL\,=\,5{,}431\,ms, WER\,=\,2.95\,\%) reaches the effective offline bound.

\paragraph{AL, DAL, and AP.}
DAL (panel~b) tracks AL closely but is consistently higher because it enforces monotone token emission; the gap narrows at larger chunk sizes as fewer out-of-order emissions occur.
AP (panel~c) rises from $0.61$ at 320\,ms to $0.80$ at $1{,}280$\,ms, reaching $0.97$ only at the $5{,}120$\,ms effective-offline point --- confirming that even the $1{,}280$\,ms operating point consumes substantially less than the full audio before emitting each token.

\paragraph{Real-time throughput.}
All six operating points run well below real-time on a single H200 GPU (maximum RTF\,=\,$0.171$ at 320\,ms chunks), confirming that \TRADE{} supports sustained streaming without GPU-throughput concerns across the entire latency range.

\end{document}